\documentclass[11pt,a4paper]{article}
\usepackage[hyperref]{emnlp-ijcnlp-2019}
\usepackage{times}
\usepackage{latexsym}
\usepackage{amsmath}
\usepackage{url}
\usepackage{graphicx}
\usepackage{color}
\usepackage{multirow}
\usepackage{array}
\usepackage{amssymb}

\aclfinalcopy 

\title{Latent Suicide Risk Detection on Microblog
via Suicide-Oriented Word \\ Embeddings and Layered Attention}

\author{Lei Cao\textsuperscript{1,2}  \and Huijun Zhang\textsuperscript{1,2}  \and Ling Feng\textsuperscript{1,2} \and Zihan Wei\textsuperscript{3} \\ 
{\bf Xin Wang}\textsuperscript{1,2} \and  {\bf Ningyun Li}\textsuperscript{1,2} \and  {\bf Xiaohao He}\textsuperscript{1,2} \\
        \textsuperscript{1}Department of Computer Science and Technology, Tsinghua University, Beijing, China \\ 
        \textsuperscript{2}Beijing National Research Center for Information Science and Technology, Beijing, China\\
        \textsuperscript{3}School of Software, Beihang University, Beijing, China \\
        {\tt \{cao-l17,zhang-hj17,xin-wang18,liny18\}@mails.tsinghua.edu.cn,}\\
        {\tt fengling@mail.tsinghua.edu.cn, sy1721114@buaa.edu.cn,}\\
        {\tt hexh17@mails.tsinghua.edu.cn} }

\date{}

\begin{document}
\maketitle
\begin{abstract}
Despite detection of suicidal ideation on social media has made great progress in recent years,
people's implicitly and anti-real contrarily expressed posts still remain as an obstacle,
constraining the detectors to acquire higher satisfactory performance.
Enlightened by the hidden ``tree holes" phenomenon on microblog,
where people at suicide risk tend to disclose their inner real feelings and thoughts
to the microblog space whose authors have committed suicide,
we explore the use of tree holes to enhance microblog-based suicide risk detection from
the following two perspectives.
(1) We build suicide-oriented word embeddings based on tree hole contents
to strength the sensibility of suicide-related lexicons and context based on tree hole contents.
(2) A two-layered attention mechanism is deployed to grasp intermittently changing points
from individual's open blog streams, revealing one's inner emotional world more or less.
Our experimental results show that with suicide-oriented word embeddings and attention,
microblog-based suicide risk detection can achieve over 91\% accuracy.
A large-scale well-labelled suicide data set is also reported in the paper.
\end{abstract}
\section{Introduction}

Suicide is a growing problem in today's society.
Each year nearly 800,000 people worldwide commit suicide,
which is one person every 40 seconds,
and there are many more who attempt it~\cite{world2014preventing}.
Suicide prevention will conduce to human's well-being, of which timely sensing suicide ideation
is an essential task.

\begin{figure}[ht]
    \centering
    \includegraphics[scale=0.42]{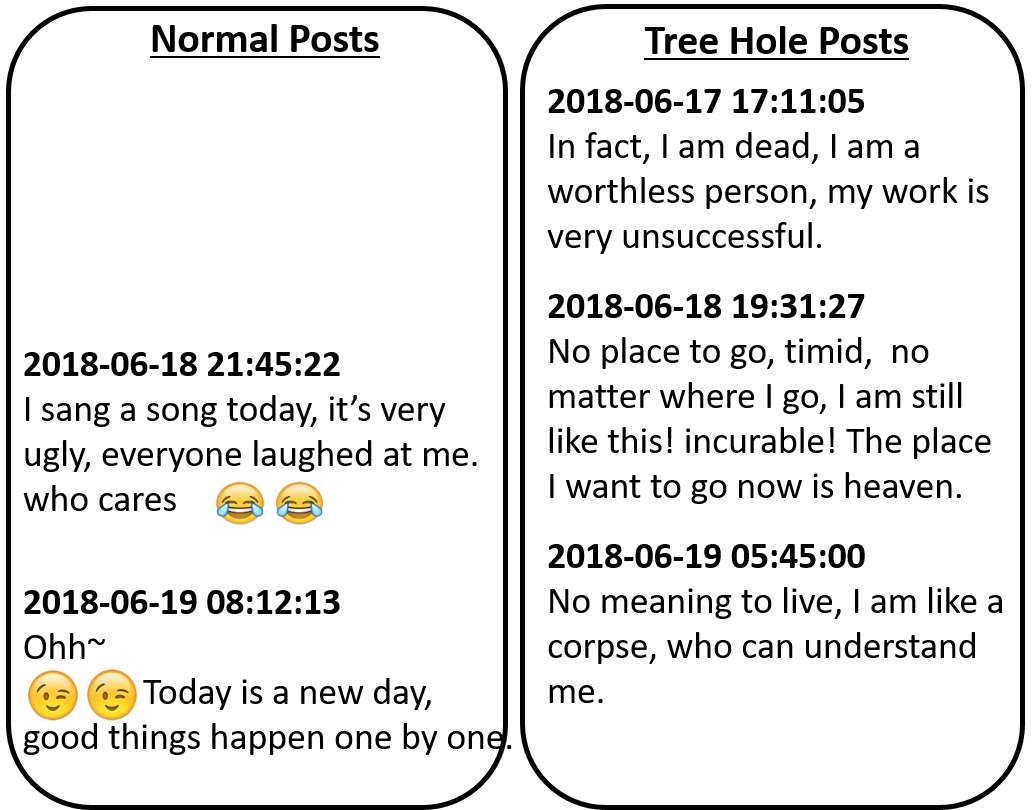}
    \caption{An example of one user's normal posts vs. his/her hidden tree hole posts on microblog.}
    \label{fig:compare}
\end{figure}

\emph{Existing Solutions.}
Traditional suicide risk assessment like
Suicide Probability Scale~\cite{bagge1998suicide},
Adult Suicide Ideation Questionnaire~\cite{fu2007predictive},
Suicidal Affect-Behavior-Cognition Scale~\cite{harris2015abc}, etc. 
requires respondents to either fill in a questionnaire or participate in a professional interview.
However, they are applicable to a small group of people.
Particularly for the people who are suffering but tend
to hide inmost thoughts and refuse to seek helps from others, the
approaches cannot function~\cite{essau2005frequency,rickwood2007and}.

\begin{table*}[t]
\centering
\begin{tabular}{m{7.5cm}|m{3.5cm}<{\centering}m{3.5cm}<{\centering}}
\hline
                  & Normal Posts  & Tree Hole Posts\\ \hline
\begin{tabular}[c]{@{}l@{}} Avg proportion of self-concern words per post   \end{tabular} &  14\%    & 50\%   \\ \hline
\begin{tabular}[c]{@{}l@{}} Avg proportion of others-concern words per post \end{tabular}&  68\%    & 12\%   \\ \hline
\begin{tabular}[c]{@{}l@{}} Avg proportion of suicide-related words per post \end{tabular}&  5\%    & 95\%   \\ \hline
\begin{tabular}[c]{@{}l@{}} Avg number of posts per user in a year\end{tabular} &  69.3    & 52.1   \\ \hline
\begin{tabular}[c]{@{}l@{}} Total number of posts from all users in a year\end{tabular} & 252,901 & 190,087 \\ \hline
\end{tabular}
\caption{Statistics of suicidal users' normal posts and hidden tree hole posts on Microblog based on 3652 users from May 1, 2018 to April 30, 2019.}
\label{table:sta}
\end{table*}

Recently, the penetration of social media (like forums and microblogs) and
its large-scale, low-cost, and open advantages
enable researchers to overcome the limitation 
and timely detect individual's suicide ideation.
Despite great efforts have been made~\cite{Alambo2019,
Cheng2017Assessing,Du2018,SawhneyMMSS18,coppersmith2018natural,Vioul2018Detection},
the social media based detection performance is constrained due to implicitly and
anti-real contrarily expressed posts from people who hide their inmost
feelings and thoughts on social media.
To illustrate, let's see a user's normal blogs vs. his/her hidden posts in a microblog tree hole
in Figure~\ref{fig:compare}.
Usually, people of suicidal tendency (referred to as suicidal people in the study) 
tend to disclose their real inner feelings on
the microblog space whose authors have committed suicide.
Hundreds of such tree holes exist on Sina microblog. 
An example tree hole contains 1,700,000 posts from suicide attempts. 
In Figure~\ref{fig:compare}, we can sense a severe hopelessness from the tree hole posts,
but not from the normal posts, and the user even expresses an uplift feeling.
After cross-examining suicidal users' normal and hidden posts as shown in table~\ref{table:sta}, 
we discover that users' hidden posts in the tree hole have more self-concerns, less others-concerns, 
and illustrate suicidal thoughts more directly.
In comparison, suicidal users normal posts contain much less suicide-related 
words, and the users are reluctant 
to show their suicidal feelings in their normal posts.
Moreover, the data of self-concern and others-concern shown in the first two rows even indicate
that people with suicide risk are not willing to talk about themselves in their normal posts,
which takes great challenges to detect suicide risk from users' open normal posts.

\emph{Our Work.}
The aim of the study is to break through the above limitation to achieve a new state-of-art 
performance on latent suicide risk detection from one's open normal microblogs.
We leverage tree hole posts from the following two perspectives.
(1) We construct suicide-oriented word embeddings based on tree hole contents
to strength the sensibility of suicide-related lexicons and context based on tree hole contents.
(2) A two-layered attention mechanism is deployed to grasp intermittently changing points
from individual's open blog streams, revealing one's inner emotional world more or less.
Our experimental results on 252,901 open normal microblogs show that 
with suicide-oriented word embeddings and two-layered attention, latent
suicide risk detection can achieve over 91\% accuracy.

In summary, the paper makes the following contributions.

\vspace{-2mm}
\begin{itemize}
\item We build effective suicide-oriented word embeddings to better understand the implicit meanings of words contained in 
users' normal posts, and propose a two-layered attention model to capture the changing points which  
reveal suicide risk from individuals' blog streams. Our latent suicide risk detection from 
users normal posts not only outperforms the state-of-the-art approaches, but also are
powerful enough in detecting implicitly and anti-real contrarily expressed posts.

\vspace{-2mm}
\item We construct a large-scale data set from 3652 suicidal people in the period of [May 1, 2018 to April 30, 2019], containing 252,901 normal posts on Sina microblog. The data set can further facilitate people's well-being studies in the computer science and psychology fields.

\end{itemize}

\section{Related Work}

\subsection{Traditional Questionnaire-based Suicide Risk Assessment}


Researchers have developed a number of psychological measurements
to access individual's suicide risk~\cite{pestian2017machine}, such as
Suicide Probability Scale (SPS)~\cite{bagge1998suicide}, Depression Anxiety Stress Scales-21 (DASS-21)~\cite{crawford2003depression,henry2005short},
Adult Suicide Ideation Questionnaire~\cite{fu2007predictive},
Suicidal Affect-Behavior-Cognition Scale~\cite{harris2015abc},
functional Magnetic Resonance Imaging (fMRI) signatures~\cite{just2017machine}, and so on.
While these measurements are professional and effective,
they require respondents to either fill in a questionnaire or participate in a professional interview,
constraining its touching to suicidal people who have low motivations to seek help from professionals~\cite{essau2005frequency,rickwood2007and,zachrisson2006utilization}.
A recent study found out that taking a suicide assessment may bring negative effect to
individuals with depressive symptoms~\cite{harris2017suicide}.

\subsection{Suicide Risk Detection from Social Media}

Recently, detection of suicide risk from social media is making great progress
due to the advantages of reaching massive population, low-cost, and real-time~\cite{braithwaite2016validating}.
\citet{harris2013} reported that suicidal users tend to spend more time online,
have greater likelihood of developing online personal relationships, and greater use of online forums.

\textbf{\emph{Suicide Risk Detection from Suicide Notes.}}
\citet{JP10} built a suicide note classifier used machine learning techniques, which performs
better than human psychologists in distinguishing fake online suicide notes from real ones.
\citet{YH07} hunted suicide notes based on lexicon-based keyword matching on MySpace.com (a popular site for adolescents and young adults, particularly sexual minority adolescents with over 1 billion registered users worldwide) to check whether users have an intent to commit suicide.

\textbf{\emph{Suicide Risk Detection from Community Forums.}}
\citet{MT13} applied textual sentiment analysis and summarization techniques to users' posts and posts' comments in a Chinese web forum in order to identify suicide expressions.
\citet{NM14} examined online forums in Japan, and discovered that
the number of communities which a user belongs to, the intransitivity,
and the fraction of suicidal neighbors in the social network
contributed the most to suicide ideation.
\citet{MC16} built a logistic regression framework to
analyze Reddit users' shift tendency from mental health sub-communities to a suicide support sub-community.
heightened self-attentional focus, poor linguistic coherence and coordination with the community, reduced social engagement and manifestation of hopelessness, anxiety, impulsiveness and loneliness in shared contents
are distinct markers characterizing these shifts.
Based on the suicide lexicons detailing
suicide indicator, suicide ideation, suicide behavior, and suicide attempt,
\citet{Alambo2019} built four corresponding semantic
clusters to group semantically similar posts on Reddit and questions in a questionnaire together, and
used the clusters to assess the aggregate suicide risk severity of a Reddit post.

\textbf{\emph{Suicide Risk Detection from Microblogs.}}
\citet{JJ14} used search keywords and phrases relevant to suicide risk factors
to filter potential suicide-related tweets,
and observed a strong correlation between
Twitter-derived suicide data and real suicide data, showing that
Twitter can be viewed as
a viable tool for real-time monitoring of suicide risk factors on a large scale.
The correlation study between suicide-related tweets and suicidal behaviors was also conducted 
based on a cross-sectional survey~\cite{HS15},
where participants answered a self-administered online questionnaire, containing questions about Twitter use, suicidal behaviour, depression and anxiety, and demographic characteristics.
The survey result showed that Twitter logs could help identify suicidal young Internet users.

Based on eight basic emotion categories
(joy, love, expectation, anxiety, sorrow, anger, hate, and surprise),
\citet{FR16} examined three accumulated emotional traits (i.e., emotion accumulation, emotion covariance,
and emotion transition) as the special statistics of emotions
expressions in blog streams for suicide risk detection.
A linear regression algorithm based on the three accumulated emotional traits was employed
to examine the relationship between emotional traits and suicide risk.
The experimental result showed that by combining all of three emotion traits together,
the proposed model could generate more discriminative suicidal prediction performance.

\begin{table*}[t]
\setlength{\belowcaptionskip}{-0.2cm}
\centering
\begin{tabular}{m{3cm}|lllll}
\hline
Category         & Suicide Ideation & Suicide behavior & Psychache & Mental illness   & Hopeless\\\hline
Number & 586 & 88 & 403 & 48 & 188\\\hline
Words/phrases & \begin{tabular}[c]{@{}l@{}}want to die\\ escape\end{tabular}  & \begin{tabular}[c]{@{}l@{}}seppuku\\ hypnotics\end{tabular} &\begin{tabular}[c]{@{}l@{}}want to cry\\ loneliness\end{tabular}&\begin{tabular}[c]{@{}l@{}}depression\\ hallucination\end{tabular}& \begin{tabular}[c]{@{}l@{}}dead end\\ despair\end{tabular}\\\hline
\end{tabular}
\caption{Representative words/phrases of Chinese suicide dictionary}
\label{table:dictionary}
\end{table*}

Natural language processing and machine learning techniques, such as
Latent Dirichlet Allocation (LDA), Logistic Regression, Random Forest,
Support Vector Machine, Naive Bayes, Decision Tree,
etc., were applied to
identify users' suicidal ideation
based on their linguistic contents and online behaviors on Sina Weibo~\cite{LG14,zhang2014using,XH14,LZ15,huang2015topic,guan2015identifying,Cheng2017Assessing}
and Twitter~\cite{Abboute2014,PB15,o2015detecting,coppersmith2015quantifying}.
Deep learning based architectures like
Convolutional Neural Network (CNN), Recurrent Neural Network (RNN), Long Short-Term Memory Neural Network (LSTM), etc.,
were also exploited to detect users' suicide risk
on social media~\cite{Du2018,SawhneyMMSS18,coppersmith2018natural}.
\citet{Vioul2018Detection}
detected users' change points in emotional well-being
on Twitter through a martingale framework, which is widely
used for change detection in data streams.





\section{Suicide-oriented Word Embeddings}
Although there are some good works on word embeddings \cite{mikolov2013efficient,pennington2014glove,joulin2016fasttext,devlin2018bert}, lack of domain information limits their performance on suicide detection.

Given a serious of pre-trained word embeddings and suicide-related dictionary, we aim to generate suicide-related word embeddings which can strengthen the sensibility of suicide-related lexicons and context.
In this study, we call them Suicide-oriented Word Embeddings, as we take advantage of the information from Tree Hole's data set which can be regarded as a kind of latent emotional expressions of individuals. 
As suicidal individuals in social media often use some suicide-related words/phrases in their posts, we employ Chinese suicide dictionary \cite{lv2015creating} to generate suicide domain associated embeddings.
The Chinese suicide dictionary analyzes 1.06 million active blog users' posts and lists 2168 words/phrases related to suicidal ideation. 
These words/phrases belong to 13 categories and each word/phrase is assigned with a risk weight from 1 to 3 which indicates the relevance of suicide. 
We list 5 representative categories in table \ref{table:dictionary}.

Since pre-trained word embeddings already contain rich semantic information and contextual information, we only need to enrich existing word embeddings with suicide-related information.

We employ a masked classification task to do this.
Generally, a sentence should contains suicide-related words/phrases if it express suicidal ideation.
Hence, We select 10,000 sentences\footnote{Through out this work, a ``sentence" can be a piece of text, rather than an actual linguistic sentence. It may contains more than one actual sentence.} from Tree Hole's data set and ensure every sentence contains more than one word/phrase appeared in Chinese suicide dictionary.

\begin{figure}
    \centering
    \includegraphics[scale=0.65]{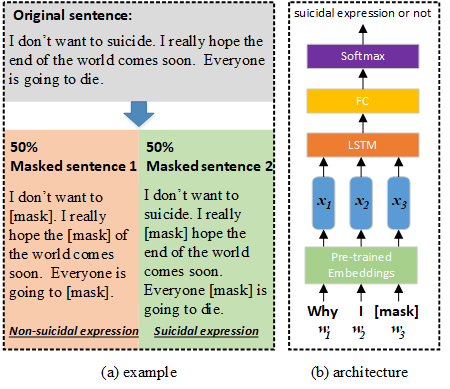}
    \caption{An example of the strategy for embeddings training is shown on the left and the architecture of the masked classification task to train the suicide-oriented word embeddings is on the right.}
    \label{fig:pre-trained}
\end{figure}

Moreover, we utilize the selected sentences to do a suicidal expression classification. A sentence is regarded as suicidal expression only if it includes at least one suicide-related word/phrase.
In this way, we do a sentence-level classification to refine pre-trained word embeddings and let them understand which word/phrase is relevant to suicide expression.

In training details, for each epoch, we randomly select 50\% sentences to replace all suicide-related words/phrases with ``[mask]".
Especially, for the rest 50\% sentences, we randomly insert two ``[mask]" into every sentence to avoid the suicidal expression classifier classifying sentence only based on whether it contains word ``[mask]".

An example is given in Figure~\ref{fig:pre-trained} (a). \textbf{Masked sentence 1} is the sentence that we replace all suicide-related words/phrases with ``[mask]" and \textbf{Masked sentence 2} is the sentence that we randomly insert two ``[mask]".
We label \textbf{Masked sentence 1} as 0 (non-suicide) and \textbf{masked sentence 2} as 1 (suicide).

As there is no clear boundary between suicide-related words/phrases and others in pre-trained word embeddings, through this suicidal expression classification we force suicide-related words/phrases to be enriched with domain information and let all the suicide-related word/phrases contain the relationship with suicidal ideation. After classification model converge in Tree Hole's data set, we obtain suicide-oriented word embeddings which contain both semantic information from pre-trained word embeddings and suicide-information from suicide dictionary.

As illustrated in Figure \ref{fig:pre-trained}, given a sentence $A = \{w_{1},w_{2},..,w_{n}\}$ written by a user in Tree Hole, where \emph{n} is the length of a sentence, the aim of suicidal expression classification is to classify  whether this sentence contains expression about suicidal ideation or not.
In this case, we define $X = \{x_{1},x_{2},..,x_{n} \} \in \mathbb{R}^{n \times d_{e}} $ as the word embeddings of \emph{A}, where $d_{e}$ is the dimension of embeddings.
Figure \ref{fig:pre-trained} shows the architecture of the suicidal expression classification Model.

We employ a LSTM layer to extract text feature from \emph{A} followed by a fully connected layer for classification. We feed the word embeddings \emph{X} into the LSTM as following:
\begin{equation}
\begin{aligned}
    &h_{t} = \textbf{LSTM}(x_{i},h_{t-1}) ,\\
    &[k_{1},k_{2}] = softmax((H^{a}W_{1} + b_{1})^{T}W_{2}+b_2)
\end{aligned}
\end{equation}
where $h_{t-1},h_{t}$ represent the hidden states at time $t-1$ and $t$, $H^{a} = \{h_{1},h_{2},...,h_{n}\} \in \mathbb{R}^{n \times d_{e}} $ is sentence representation of \emph{A}, $[k_{1},k_{2}]$ stand for the possibility of whether the sentence contains expression about suicidal ideation or not.  
$W_{1} \in \mathbb{R}^{d_{e} \times 1}$, $W_{2} \in \mathbb{R}^{n \times 2}$, $b_{1} \in \mathbb{R}^{1 \times 1}$ and $b_{2} \in \mathbb{R}^{1 \times 2}$ are trainable parameters.

\section{Suicide Detection Model (SDM) based on Suicide-oriented Word Embeddings and Attention Mechanism}

\begin{figure}
    \centering
    \includegraphics[scale=0.53]{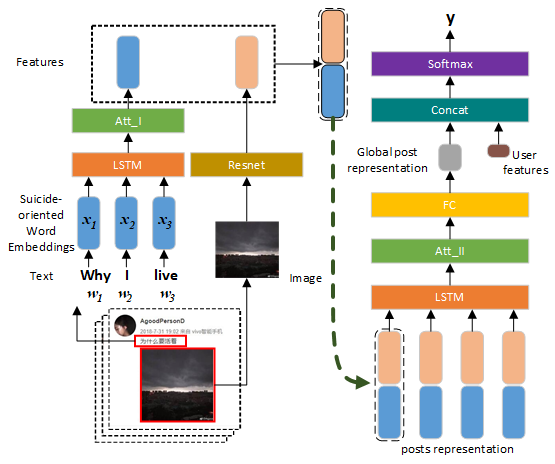}
    \caption{Architecture of the Suicide Risk Detection model.}
    \label{fig:model}
\end{figure}

Given a sequential of posts $\hat{T}$ from one user, $\hat{T} = \{(s_{1},p_{1}),(s_{2},p_{2}),...,(s_{m},p_{m})\}$ , where \emph{m} denotes the number of posts, $(s_{i},p_{i})$ stand for text and picture from i-th post. 
The aim is to detect whether the user at risk of suicide or not.
Let $\hat{X} = \{x_{1},x_{2},..,x_{n} \} \in \mathbb{R}^{n \times d_{e}} $ be the word embeddings of $s_{i}$, where \emph{n} represents the length of $s_{i}$ and $d_{e}$ is the dimension of embeddings.
Figure \ref{fig:model} shows the architecture of the proposed two-layered attention model.
\subsection{Feature Extraction}
\noindent\textbf{Text Feature Extraction.}
We employ a LSTM layer and attention mechanism to extract text feature from $s_{i}$. We feed the word embeddings $\hat{X}$ into the LSTM as following:
\begin{equation}
h_t =\textbf{LSTM}(x_t, h_{t-1})
\end{equation}
where $h_{t-1},h_{t}$ represent the hidden states at time $t-1$ and $t$.
We obtain the primary textual representation $H_{i}^{p} = \{h_{1},h_{2},...,h_{n}\} \in \mathbb{R}^{n \times d_{e}} $ of $s_{i}$ after LSTM layer.
To gain the critical suicide-related textual information of $H_{i}^{p}$, we apply the \textbf{attention mechanism \textit{Att\_I}} :
\begin{equation}
Att\_I = softmax(H_{i}^{p}W_{3}+ b_{3})
\end{equation}

where $Att\_I \in \mathbb{R}^{n \times 1} $ is the attention vector  that  demonstrates the distribution of the  weights for each word of primary textual representation, $W_{3} \in \mathbb{R}^{d_e \times 1}$ and $b_{3} \in \mathbb{R}^{1 \times 1}$ are trainable parameters.
Then we make multiplication between the attention vector $Att\_I$ and $H_i^p$ to get the final textual representation $\hat{H_{i}} \in \mathbb{R}^{d_{e}\times 1}$ of text $s_{i}$,
\begin{equation}
\hat{H_{i}} = (H_{i}^{p})^{T} Att\_I 
\end{equation}

\begin{table*}[t]
\begin{tabular}{l|c|l}
\hline
Feature            & Dimension & Description                            \\ \hline\hline
Gender                  & 3   & (1,0,0)for male, (0,1,0)for female and (0,0,1) for `Unknown'.    \\\hline
Screen name length     & 1     & The length of screen name.                                        \\\hline
Post Count             & 1     & The number of posts.                                      \\\hline
Follower              & 1         & The number of followers.                                      \\\hline
Following               & 1     & The number of following.                                      \\\hline
Picture             & 1         & The number of posts with picture.                                       \\ \hline
Post time      & 4           & \begin{tabular}[c]{@{}l@{}}The four dimensions are proportions of posts posted in \\ (0:00-5:59), (6:00-11:59), (12:00-17:59), (18:00-23:59) of a day.\end{tabular}                                        \\ \hline
\end{tabular}
\caption{Summary of user's features.}
\label{table:feature}
\end{table*}

\noindent\textbf{Image Feature Extraction.}
We extract image features from a 34 layer pre-trained ResNet \cite{he2016deep}. For the convenience of calculation, we convert the last fully connected layer of ResNet from $512 \times 1000$  to $512 \times d_{e}$:
\vspace{-2mm}
\begin{equation}
I_{i} = tanh(O~W_{4}+b_4)
\end{equation}
where  $O \in \mathbb{R}^{1 \times 512}$ is the input of the last fc-layer, $W_{4} \in \mathbb{R}^{512 \times d_{e}}$ and $b_{4} \in \mathbb{R}^{1 \times d_{e}}$ are trainable parameters.
Then, $I_{i} \in \mathbb{R}^{1 \times d_{e}}$ is the visual representation of picture $p_{i}$.

\noindent\textbf{User's Feature Extraction.}

As illustrated in table \ref{table:feature}, we extract 12 features $F \in \mathbb{R}^{12 \times 1}$ from user' profile and posting behaviour. Since not every one has tree hole's data, in this study we do not consider tree hole's data in suicide risk detection model.

\subsection{Suicide Risk Detection}

Given textual representation $\hat{H_{i}}$ and visual representation $I_{i}$, we employ a concatenate operation $\oplus$ to obtain the post representation $E_{i} \in \mathbb{R}^{2d_{e} \times 1}$ for a single post $(s_{i},p_{i}) $:
\begin{equation}
E_{i} = \hat{H}_{i} \oplus I_{i}^{T}
\end{equation}

Similar with above, we employ another LSTM layer and attention mechanism to generate global post representation $G \in \mathbb{R}^{30\times 1}$:
\begin{equation}
\begin{aligned}
    &h_{t}= \textbf{LSTM}(E_{i},h_{t-1}) ,\\
    &Att\_II = softmax(H^{g}W_{5}+b_{5}),\\
    &G = tanh(((Att\_II)^{T} \times H^{g})W_{6} + b_{6}),
\end{aligned}
\end{equation}

where $h_{t-1},h_{t}$ represent the hidden states at time $t-1$ and $t$.
$H^{g} = \{h_{1},h_{2},...,h_{m}\} \in \mathbb{R}^{m \times d_{e}} $ represents the primary post representation of a user after LSTM layer.
As not every post of a user expresses the ideation of suicide,  we apply the \textbf{attention mechanism \textit{Att\_II}} to gain the high suicide risk post information of $H^{e}$.
An attention vector $Att\_II \in \mathbb{R}^{m \times 1}$ was computed to present the different risk weight of posts, where $W_{5} \in \mathbb{R}^{d_{e} \times 1}$ and $b_{5} \in \mathbb{R}^{1 \times 1}$ stand for trainable parameters.
Then based on attention $Att\_II$, we obtain the global post representation $G$ for a user, where $W_{6} \in \mathbb{R}^{d_{e} \times 30}$ and $b_{6} \in \mathbb{R}^{1 \times 30}$ stand for trainable parameters.

Finally, we apply a concatenate operation to jointly consider $G$ and $F$, and through a fully connected layer to compute the possibility of suicide:
\begin{equation}
[y_1,y_0] = softmax(( G \oplus F)^{T}W_{7} +b_{7}) 
\end{equation}

where $y_{1},y_{0}$ represent the possibility of a user at risk of suicide or not, $W_{7} \in \mathbb{R}^{42 \times 2}$ and $b_{7} \in \mathbb{R}^{1 \times 2}$ stand for trainable parameters.

\section{Experiments}

\begin{table}[t]
\setlength{\belowcaptionskip}{-0.3cm}
\centering
\begin{tabular}{c|c|c|m{2cm}<{\centering}}
\hline
                   & Users     & Posts  &\begin{tabular}[c]{@{}l@{}} Posts with \\ image\end{tabular}  \\ \hline
  suicide            & 3,652 & 252,901 & 93,461 \\
  non-suicide            & 3,677  &491,130    &  260,667    \\\hline
\end{tabular}
\caption{Statistic of suicide data set.}
\label{table:sta-d2d3}
\end{table}

\subsection{Data Collection}
To make suicide risk detection via social media, we construct two data sets: one from Tree Hole and another from Weibo.

\textbf{Tree Hole's data set.}
We studied a suicidal community which exists in the comments of a Chinese student's last posting before the student committed suicide.
In March 17, 2012, this student which screen name is ``Zoufan" left the last word on Weibo and then committed suicide. For the past seven years, More than 160,000 people gather here and write over 1,700,000 comments which is still continuing to grow. They express their suicidal thoughts, show their tragic experiences and demonstrate their plans of suicide behaviors. In psychology, we can understand this community as a Tree Hole.
We crawled all comments from May 1,2018 to April 30, 2019 and selected top 4,000 active users. After that, four doctoral students major in computational mental healthcare were employed to annotate users that whether they are at risk of suicide or not.
Specifically, We only decide the user "at suicide risk" based on self-report of his/her tree hole posts.
If a user express clear suicidal thoughts like \emph{``At this moment, I especially want to die. I feel very tired. I really want to be free."} more than 5 times in different days, then we label him/her at suicide risk.
Finally we get 190,087 sentences of 3,652 users and the average length per post is 11.96 words.

\textbf{Suicide data set.}
To collect users at suicide risk, we crawl user's profile and all created posts in Weibo according to user list from Tree Hole' data set. Besides, we select the users who never submit any post containing expression about suicidal ideation and label them as non-suicide risk. 
In this case, we discard users whose fans more than 1,500 or posts more than 2,000 because that they may be public figure or organization. 
The statistic of suicide data set are illustrated in Table \ref{table:sta-d2d3}.

\begin{table*}[t]
\begin{tabular}{c|c|c|c|c|c|c|c|c}
\hline
\multicolumn{2}{l|}{}                                  & Word2vec & GloVe & FastText & Bert & So-W2v & So-GloVe & So-FastText                     \\ \hline
\multicolumn{1}{l|}{\multirow{2}{*}{LSTM}} & Acc(\%) & 79.21    & 80.17      & 82.59    & 85.15      & 86.00       & 86.45         & \textbf{88.00} \\ \cline{2-9} 
\multicolumn{1}{l|}{}                      & F1(\%) & 78.58    & 79.98       & 82.18    & 85.69     & 86.17       & 86.69         & \textbf{88.14} \\ \hline
\multicolumn{1}{l|}{\multirow{2}{*}{SDM}} & Acc(\%) & 86.54    & 86.55      & 87.08    & 88.89     & 90.83       &  91.00        & \textbf{91.33} \\ \cline{2-9} 
\multicolumn{1}{l|}{}                      & F1(\%) & 86.63    & 85.13      & 86.91    & 87.44     & 90.55       &  90.56        & \textbf{90.92} \\ \hline
\end{tabular}
\caption{Performance comparison for different word embedding and different detection model, where ``So-W2v", ``So-Glove" and ``So-FastText" represent suicide-oriented word embeddings based on
Word2vec, GloVe and FastText respectively. Acc and F1 represent accuracy and F1-score.}
\label{table:performance}
\end{table*}


\begin{table}[t]
\centering
\begin{tabular}{c|c|c|c|c}
\hline
     & \multicolumn{2}{c|}{\begin{tabular}[c]{@{}c@{}}Full testset\end{tabular}} & \multicolumn{2}{c}{\begin{tabular}[c]{@{}c@{}}Harder sub-testset\end{tabular}} \\ \hline
     & Acc                                  & F1                            & Acc                               & F1                              \\ \hline
SVM  & 70.34                                 & 69.01                                & 61.17                                    & 64.11                                   \\ \hline
NB   & 69.59                                 & 70.12                                & 65.14                                    & 62.20                                   \\ \hline
LSTM & 88.00                                 & 88.14                                & 76.89                                    & 75.32                                   \\ \hline
SDM & \textbf{91.33}                        & \textbf{90.92}                    & \textbf{85.51}                              & \textbf{84.77}                          \\ \hline
\end{tabular}
\caption{Performance comparison between different suicide risk detection model, where Acc and F1 represent accuracy and F1-score respectively. }
\label{tab:compare_model}
\end{table}

\subsection{Data Preprocessing}
We carry out the following data preprocessing procedures: 
1) \textbf{Emoji}. We replace emoji with corresponding word like ``happy", ``cry" to facilitate our model to understand the emotion of user's post.
2) \textbf{URL.} As URL has no use for our detection, we simply remove them from sentences.
3) \textbf{Image.} All images posted by users were adjusted to $224 \times 224$ for normalized input.
\subsection{Experimental Setup}
In suicide detection task, we treat recent 100 posts from one user as one sample. 
After sum up $D_2$ and $D_3$, we obtain 7,329 microblog users and training set, validation set and test set contain 6,129, 600, 600 respectively.
All sentences are padded to the length of the longest sentence in the data set with word ``$<$PAD$>$". 
Batch size is 16 during training process and we use 0.001 as learning rate. Adam \citet{Kingma2015Adam} is adopted as the optimizer.

\begin{figure*}[ht]
	\centering
		\includegraphics[width=0.5\columnwidth]{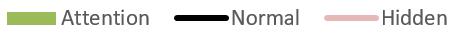}
	\begin{tabular}{cccc}		
		\includegraphics[width=0.46\columnwidth]{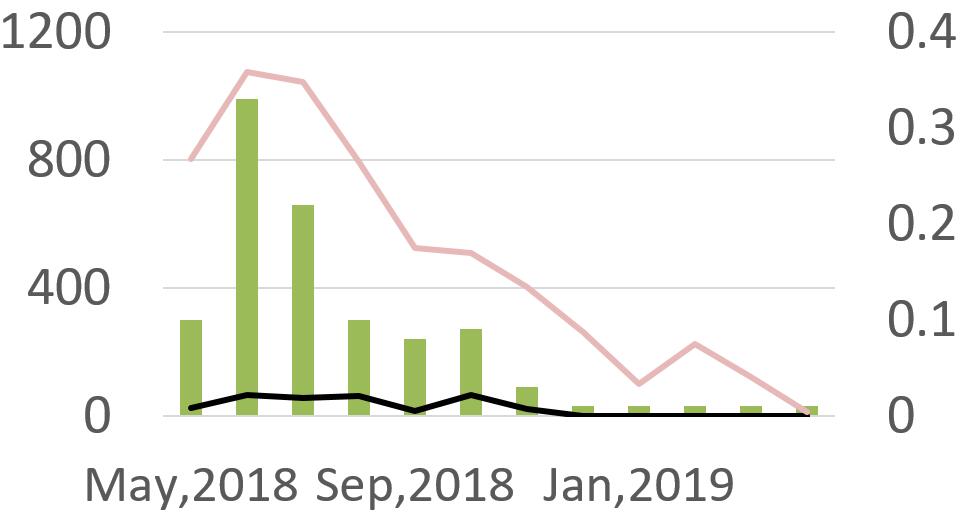} &
		\includegraphics[width=0.46\columnwidth]{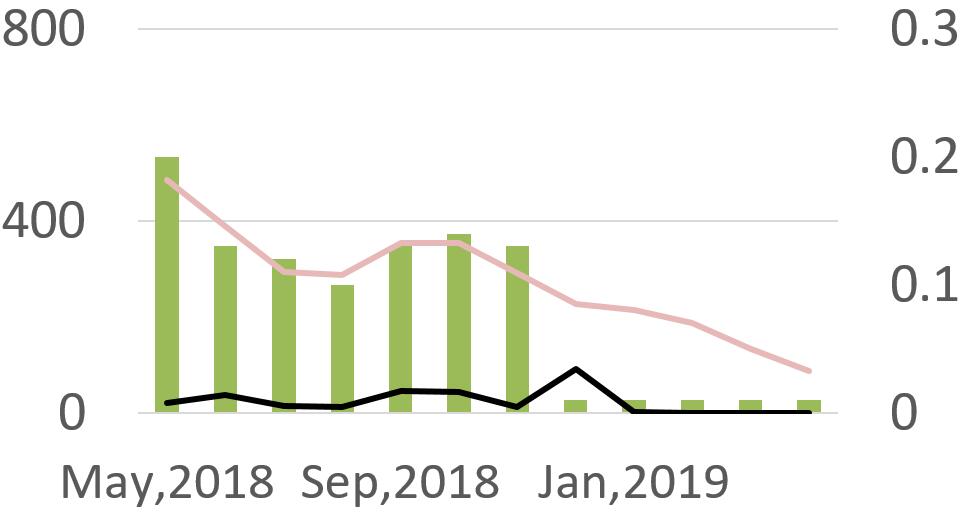} & 
		\includegraphics[width=0.46\columnwidth]{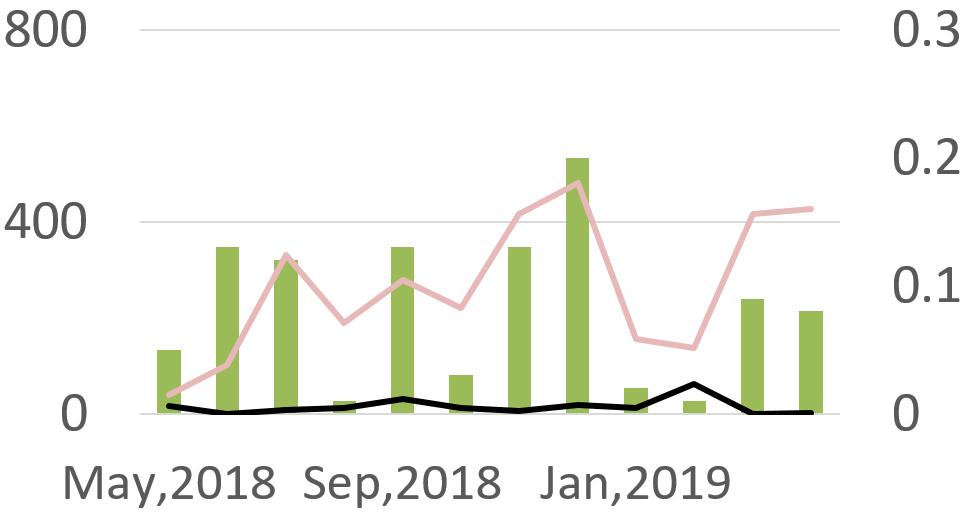} &
		\includegraphics[width=0.46\columnwidth]{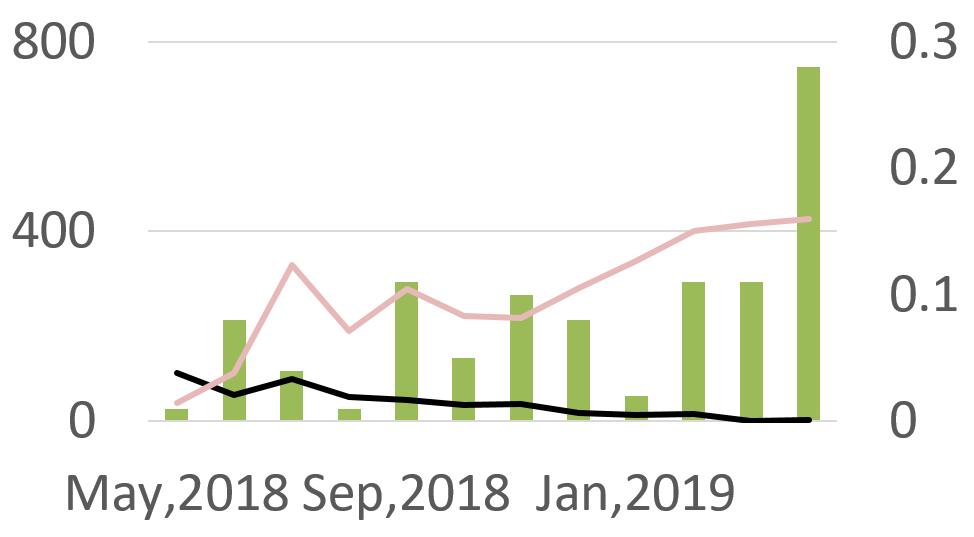}  \\
		{\small\begin{tabular}[c]{@{}l@{}} (a)  \textsl{User 1 with $\rho_{nh}=0.84$} \\  \textsl{and  $\rho_{ah}=0.88$}   \end{tabular}} 
		&  {\small\begin{tabular}[c]{@{}l@{}} (b)  \textsl{User 2 with $\rho_{nh}=0.34$} \\  \textsl{and  $\rho_{ah}=0.91$}   \end{tabular}} 
		& {\small\begin{tabular}[c]{@{}l@{}} (c)  \textsl{User 3 with $\rho_{nh}=-0.12$} \\  \textsl{and  $\rho_{ah}=0.6$}   \end{tabular}} 
		&  {\small\begin{tabular}[c]{@{}l@{}} (d)  \textsl{User 4 with} \\ \textsl{  $\rho_{nh}= -0.7$ and  $\rho_{ah}=0.6$}   \end{tabular}}
	\end{tabular}

	\caption{ Correlation between normal posts, hidden posts and attention on four representative suicide risk users. The x-axis shows dates from May, 2018 to April, 2019. The left y-axis represent number of suicde-related words/phrases and the right represent attention weight. }
	\label{fig:pearson}
\end{figure*}

We compare suicide-oriented word embeddings with following well-developed word embeddings.

(1) \textbf{Word2vec}: The fundamental work for considering the local semantic information of words. We get pre-trained Word2vec word embeddings from \citet{li2018analogical}.

(2) \textbf{GloVe}: Context-based unsupervised algorithm, which apply co-occurrence matrix to jointly consider  the  local  and  global  semantic  information. We apply open source tool \footnote{https://github.com/stanfordnlp/GloVe} to train word embeddings on all sentences from Tree Hole's data set. 

(3) \textbf{Fasttext}: A fast text classification and representation learning model based on Word2vec and hierarchical softmax. Pre-trained FastText word embeddings were obtained from official project \footnote{https://github.com/facebookresearch/fastText}.

(4) \textbf{Bert}: Latest language model based on transformer. We acquire pre-trained Bert model from official project\footnote{https://github.com/google-research/bert} and generate word embedding of each word in a sentence dynamically.

Also, we compare our suicide risk detection model with following well-designed methods.

(1) \textbf{LSTM} \cite{coppersmith2018natural}: An attention mechanism based Long Short-Term Memory model which can capture  contextual information between suicide-related words and others.

(2) \textbf{Naive Bayesian (NB) and Support Vector Machine (SVM)} \cite{pedregosa2011scikit}: Two representative machine learning methods with well-designed features. We use SC-LIWC information \cite{Cheng2017Assessing} as textual features, saturation, brightness, warm/clear color and five-color theme information \cite{shen2018cross} from picture as visual features and user's behaviour features from table \ref{table:feature}.


\subsection{Results}
Three sets of tests were conducted to evaluate the performance of suicide risk detection model with suicide-oriented word embeddings.
\subsubsection{Effectiveness of Suicide-oriented Word Embeddings}
We compare the performance of LSTM and SDM with seven word embeddings as illustrated in table \ref{table:performance}. We find that without suicide-related dictionary, Bert outperforms other three word embeddings with 2\% higher accuracy and 1.5\% higher F1-score on both models. After leveraging suicide-related dictionary, suicide-oriented word embeddings based on FastText achieves the best performance with accuracy 88.00\% 91.33\%, F1-score 88.14\%, 90.92\% on two models. Obviously, there is a gap between suicide-oriented word embeddings and normal word embeddings which can verify the effectiveness of the former.
\subsubsection{Effectiveness of Suicide Risk Detection Model}
We compare the performance of four model as shown in table \ref{tab:compare_model}. In this case, LSTM and SDM employs So-FastText word embeddings as their input.
SDM improves the accuracy by over 3.33\% and obtains 2.78\% higher F1-score on full data set.
\subsubsection{Performance on \textsl{Harder sub-testset}}
To verify the effectiveness of models in dealing with people’s implicit and anti-real contrary expressions on microblog posts, we filter out 130 suicide risk people from test set who do not show obvious suicidal ideation on normal posts. Those 130 people construct a subset of test set named \textsl{Harder sub-testset}. After observing the performance of four models as shown in table \ref{tab:compare_model}, SDM can keep 8\% higher value both in accuracy and F1-score on
\textsl{Harder sub-testset}, compared with other models, 
and the decline is smaller than other models.
This suggests that SDM performs better than existing models in dealing with people's implicit and anti-real contrary expressions.

\subsubsection{Ablation Test for Suicide Risk Detection Model}
\begin{table}[]
\begin{tabular}{l|c|c}
\hline
Inputs                                                                & Accuracy & F1-score \\ \hline
Text                                                                 & 88.56    & 87.99    \\ \hline
Text+Image                                                           & 89.22    & 89.22    \\ \hline
Text+User's feature                                                  & 90.66    & 90.17    \\ \hline
\begin{tabular}[c]{@{}l@{}}Text+Image\\ +User's feature\end{tabular} & \textbf{91.33}    & \textbf{90.92}    \\ \hline
\end{tabular}
\caption{Ablation test for SDM with different inputs.}
\label{tab:ablation}
\end{table}

To show the contribution of different input to the final classification performance, we design an ablation test of SDM after removing different input.
All SDMs are based on embedding So-Fasttext. 
Since not every post contains image and user's features contain missing value, we do not only use images nor user's feature as input of SDMs. 
As illustrated in table \ref{tab:ablation}, we can see that textual information is a crucial input of our SDM.
Besides, user's features play a more important role than visual information.
The more modalities we use, the better performance we get. 
\subsection{Discovery}
To further explore the correlation between normal posts and hidden posts from same user, we import \textsl{Pearson Correlation Coefficient} \cite{tutorials2014pearson} to manage it.
For each user, we obtain a normal vector $V_{i}^{n} = \{v_{i,Jan}^{n},v_{i,Feb}^{n},...,v_{i,Dec}^{n}\} \in \mathbb{R}^{12}$, where $v_{i,Jan}^{n}$ shows the total number of times words/phrases appear in posts for user \emph{i} in January. In a similar way we get a hidden vector $V_{i}^{h} = \{v_{i,Jan}^{h},v_{i,Feb}^{h},...,v_{i,Dec}^{h}\} \in \mathbb{R}^{12}$.
 For each normal posts, we also have an attention weight from $Att_{II}$ which represents the suicide risk. Then, similar as above, an attention risk vector $V_{i}^{a} = \{v_{i,Jan}^{a},v_{i,Feb}^{a},...,v_{i,Dec}^{a}\} \in \mathbb{R}^{12}$ was computed, where $v_{i,Jan}^{a}$ shows the total suicide risk for user \emph{i} in January.
 We donate the pearson correlation coefficient $\rho_{n,h,i}$ of $V_{i}^{n}$ and $V_{i}^{h}$, $\rho_{a,h,i}$ of $V_{i}^{a}$ and $V_{i}^{h}$  as the correlation between normal posts and hidden posts , attention and hidden posts for user \emph{i}.

As shown in figure \ref{fig:pearson}, we find that there are high positive linear correlation between normal posts and hidden posts from user 1 with $\rho_{nh} = 0.84$ and high negative linear correlation
from user 4 with $\rho_{nh} = -0.7$. For other two suicide risk users, there are not obvious linear correlations with $\rho_{nh} = 0.33, -0.12$ respectively.
The phenomenon that correlations $\rho_{ah}$ between attention and hidden posts from four users all higher than 0.6 which means high positive linear correlation, which verify the ability of the two-layered attention mechanism to reveal ones' inner emotional world.

\section{Conclusion}
In this paper, we explore the uses of tree holes to enhance microblog-based suicide risk detection.
Suicide-oriented word embeddings based on tree hole contents are built to strengthen the sensibility of suicide-related lexicons and a two-layered attention mechanism is deployed to grasp intermittently changing points from individual’s open blog streams.
Based on above word embeddings and attention mechanism, we propose a suicide risk detection model which outperforms the well-designed approaches on benchmark data set.
Through experimental results we also find that, our model also performs well on people's implicit and anti-real contrary expressions.

\section*{Acknowledgments}
The work is supported by National Natural Science Foundation of China (61872214, 61532015, 61521002) and Chinese Major State Basic Research Development 973 Program (2015CB352301).

\bibliography{suicide}
\bibliographystyle{acl_natbib}
\end{document}